\pdfoutput=1

\documentclass[11pt]{article}

\usepackage[preprint]{acl}

\usepackage{times}
\usepackage{latexsym}

\usepackage[T1]{fontenc}

\usepackage[utf8]{inputenc}

\usepackage{microtype}

\usepackage{inconsolata}

\usepackage{graphicx}

\usepackage{multirow}
\usepackage{enumitem}
\usepackage{xcolor,colortbl}
\usepackage{soul}
\sethlcolor{yellow}
\usepackage{tabularx}
\usepackage{tcolorbox}
\usepackage{comment}
\usepackage{amsmath}  
\usepackage{booktabs}
\usepackage{breqn}
\usepackage{soul}
\usepackage{xcolor,colortbl} 
\usepackage{changepage,threeparttable} 
\usepackage{adjustbox}
\usepackage{lineno}
\usepackage{makecell}
\title{Decision-Oriented Text Evaluation}

\author{
 \textbf{Yu-Shiang Huang,\textsuperscript{1, 2}}
 \textbf{Chuan-Ju Wang,\textsuperscript{1}}
 \textbf{Chung-Chi Chen\textsuperscript{3}}
\\
 \textsuperscript{1}Academia Sinica, Taiwan \\
 \textsuperscript{2}National Taiwan University, Taiwan\\
 \textsuperscript{3}National Institute of Advanced Industrial Science and Technology, Japan
\\
 \texttt{F09946004@ntu.edu.tw, cjwang@citi.sinica.edu.tw, c.c.chen@acm.org}
}

\begin{document}
\maketitle


\begin{abstract}
Natural language generation (NLG) is increasingly deployed in high-stakes domains, yet common intrinsic evaluation methods, such as n-gram overlap or sentence plausibility, weakly correlate with actual decision-making efficacy. We propose a decision-oriented framework for evaluating generated text by directly measuring its influence on human and large language model (LLM) decision outcomes. Using market digest texts---including objective morning summaries and subjective closing-bell analyses---as test cases, we assess decision quality based on the financial performance of trades executed by human investors and autonomous LLM agents informed exclusively by these texts. 
Our findings reveal that neither humans nor LLM agents consistently surpass random performance when relying solely on summaries. However, richer analytical commentaries enable collaborative human–LLM teams to outperform individual human or agent baselines significantly. Our approach underscores the importance of evaluating generated text by its ability to facilitate synergistic decision-making between humans and LLMs, highlighting critical limitations of traditional intrinsic metrics.
\end{abstract}

\section{Introduction}

Natural language generation (NLG) systems are rapidly transitioning from research environments to high-stakes domains such as finance, medicine, and law, where the text they produce can influence decisions involving money, health, or liberty.  
Nonetheless, mainstream evaluation continues to emphasize intrinsic measures such as $n$-gram overlap (\textsc{BLEU}, \textsc{ROUGE}) or sentence-level plausibility scores, despite substantial evidence that these metrics correlate only weakly with downstream utility or reliability \cite{callison-burch2006evaluating, novikova-etal-2017-need,pagnoni-etal-2021-understanding}.  
A summary may appear ``faithful'' while omitting a material risk; a fluent discharge note may lead a clinician toward an unsafe treatment; a well-phrased market brief may guide investors into loss-making positions.  
The key question, therefore, becomes: How should we assess generated text when its value is ultimately determined by the decisions it informs?

We respond by adopting a decision-oriented perspective on evaluation.  
Rather than evaluating whether a system mirrors reference phrasing or achieves high mean-opinion scores, we assess whether its output enables human or automated agents to make sound decisions.  
Our test bed is daily market digest, a genre with immediate and quantifiable economic implications.  
Specifically, we examine (i) morning briefs, which concisely summarise the previous session and highlight salient overnight events, an essentially objective summarisation task, and (ii) closing-bell reports, which interpret intraday movements and speculate about the next session, an inherently subjective analytical task.  
We evaluate their quality by the buy/sell performance of (a) retail investors and (b) frontier-scale LLM agents trading solely based on the provided text.

Previous efforts toward aligning summarization evaluation with decision-making outcomes, such as decision-focused summarization \cite{hsu-tan-2021-decision} and extrinsic evaluation through QA or classification tasks \cite{pu-etal-2024-summary}, largely overlook the significant analytical capacities that modern LLMs inherently possess. Contemporary LLMs can generate content that extends beyond summarization into meaningful analytical commentary, directly informing effective decisions. This paper contributes to an emerging discourse by demonstrating that LLM-generated content, whether objective or subjective, significantly influences decision-making efficacy and can notably enhance the performance of human–LLM collaborative teams.
In our experiments, we find distinct patterns depending on the report type: For morning briefs, neither human nor LLM investors achieve optimal decision performance when relying solely on human-authored content, whereas LLM-generated summaries consistently enhance decision accuracy. In contrast, for closing-bell reports, human-generated texts significantly improve decision quality for LLM investors, highlighting the complementary strengths of human analysis and machine-generated insights. Notably, detailed analytical commentaries foster effective collaboration between human and LLM investors. These results indicate the inadequacy of traditional surface-level metrics and advocate for decision-oriented evaluation methods, emphasizing the practical, real-world value of generated texts. Our work aims to establish a foundation for systematically assessing and enhancing NLG outputs in high-stakes decision-making scenarios.


\section{Related Work}

Decision‐oriented summarization has shifted focus from surface metrics to downstream impact. In healthcare, \citet{hsu-tan-2021-decision} proposes decision faithfulness metrics to gauge whether summaries support correct clinical actions. In finance, \citet{takayanagi-etal-2025-gpt} find that GPT-4–generated reports can sway investor choices, especially among novices. 
These studies underscore the importance of assessing real-world impact, yet they often focus on classification-style tasks or single-user decision accuracy, without closely examining how generated text affects actual downstream decision behavior in complex settings.
In high‐stakes fields like finance and medicine, misleading summaries can have severe consequences, motivating outcome‐based evaluation frameworks \citep{zhang-etal-2020-medical, joshi-etal-2022-medical}. 
Building on this, we introduce a consequence‐driven evaluation tailored to market commentaries, systematically measuring how human and LLM narratives affect actual investment decisions by both people and autonomous agents.

\begin{table*}[t]
  \centering
  \resizebox{\textwidth}{!}{
    \begin{tabular}{ll|r|rr|r|rr}
    \multicolumn{2}{c|}{\multirow{2}[1]{*}{Investor}} & \multicolumn{3}{c|}{Morning Briefs} & \multicolumn{3}{c}{ Closing-Bell Reports} \\
    \multicolumn{2}{c|}{} & \multicolumn{1}{c|}{Journalist} & \multicolumn{1}{c}{Performance-Based} & \multicolumn{1}{c|}{Professional-Insight} & \multicolumn{1}{c|}{Journalist} & \multicolumn{1}{c}{Performance-Based} & \multicolumn{1}{c}{Professional-Insight} \\
    \hline
    \multirow{3}[2]{*}{LLM} & Claude-3-5-Sonnet & 38.85 & 45.98 & \textbf{48.01} & \textbf{65.56} & 55.62 & 60.07 \\
          & Gemini-2.0-Flash & 44.89 & \textbf{46.98} & 42.41 & \textbf{61.60} & 54.36 & 58.25 \\
          & GPT-4o & 42.35 & 42.53 & \textbf{43.15} & \textbf{58.51} & 56.89 & 55.17 \\
    \hline
    \multirow{3}[1]{*}{Human} & A     & 39.64 & 43.10 & \textbf{48.61} & 47.59 & 48.97 & \textbf{56.71} \\
          & B     & 36.67 & \textbf{45.11} & 40.23 & 50.83 & 49.44 & \textbf{53.45} \\
          & C     & 34.40 & \textbf{49.27} & 48.35 & 42.24 & \textbf{75.00} & 54.18 \\
    \end{tabular}%
    }
  \caption{\textbf{The accuracy of investor decisions under different market digests (\%).} Journalist refers to verbatim transcripts from professional financial news channels. The bolded results indicate the best performance of the same investor under identical decision-making scenarios.}
  \label{tab:accuracy}%
  \vspace{-0.4cm}
\end{table*}%

\section{Experimental Design}

\subsection{Dataset and Market Digest Generation}
We target two key forms of financial commentary: morning briefs—daily summaries of the previous trading day—and closing‐bell reports—analytical end‐of‐day commentaries informed by intraday performance.\footnote{An illustrative example appears in Appendix~\ref{app:example}.}  
To build our dataset, we collected verbatim transcripts from professional financial news channels over a 30-day window, aligning each brief and report with its corresponding market events.  On average, each day’s narratives drew from roughly 2,400 distinct news articles, with closing‐bell reports further enriched by expert analysis and intraday statistics.

To assess how LLM‐generated commentaries influence decision‐making relative to expert texts, we used GPT-4o to synthesize morning briefs and closing‐bell reports under various prompt configurations.  
Facing the impracticality of supplying the full set of over 2,000 articles at once, we employed two‐stage pipelines—first selecting a subset of salient assets, then generating the market digest conditioned on that selection.  

First, we implement a \textbf{performance-based} selection pipeline that drives both morning and closing briefs. 
For each trading day, we rank the prior day’s equities by three metrics---price volatility, trading volume, and net institutional buy-sell imbalance---and select the top $K$ performers in each category. The news about these high-momentum, high-liquidity stocks forms the candidate set for the morning brief, where the LLM generates a concise summary of overnight developments. 
To generate the closing‐bell report, we provide the same day’s generated morning brief along with that day’s top $K$ intraday data—price changes, trading volumes, and net institutional flows—as input, enabling the model to deliver a truly market‐anchored end‐of‐day analysis.  

Second, our \textbf{professional‐insight} selection pipeline isolates the effect of expert curation. We first employ GPT-4o to extract every company explicitly mentioned in the professional human transcript, then retrieve the corresponding news articles for that curated list. This human-selected set serves as the input for the morning‐brief and closing‐bell generation. By generating narratives exclusively around these professionally highlighted equities, we can directly measure how expert‐driven stock selection influences commentary coherence and predictive accuracy.  


\subsection{Decision-oriented Evaluation}
Our preliminary observations indicated that LLMs can produce fluent and coherent morning briefs and closing-bell reports, yet these generated texts significantly differ from their professionally authored counterparts. 
Consequently, pursuing complete lexical alignment with expert-produced texts may not be a meaningful evaluation criterion. 
Instead, we propose treating LLM-generated texts as independent analytical contributions, warranting evaluation through the same decision-oriented lens used to assess human-generated commentary.

Following this rationale, the primary purpose of market digests is to guide informed investment decisions. 
To rigorously isolate the impact of the generated texts on decision-making outcomes, we constrained our evaluation scenarios within very short time horizons, thereby minimizing confounding factors such as broader market trends or external events.
Participants—both human investors and LLM agents—used each market digest to select only the stocks they felt confident would move by market close (for morning briefs) or by the next day’s open (for closing‐bell reports). 
Importantly, participants were not required to forecast the movements of all stocks; instead, they selected only those assets they felt confident enough to trade. It is a realistic reflection of actual investor behavior.
We employ thresholded prediction accuracy, which labels returns above $+0.55\%$ as “rise” and below $-0.50\%$ as “fall”~\cite{xu-cohen-2018-stock}, as our primary metric for market digest quality, directly quantifying how different text‐generation strategies influence real-world financial decision-making.

In our evaluation, we employed three LLMs as agent investors (GPT-4o, Gemini-2.0-Flash, and Claude-3.5-Sonnet) and invited three experienced human investors to participate in the experiment. To ensure fairness, participants were prohibited from accessing any external reference materials. The basic hourly wage is set at 130\% of the statutory minimum wage. To incentivize participation, we implemented the following bonus scheme: if an annotator's overall investment success rate exceeded that of the LLMs, they received an additional bonus of USD 65. Furthermore, the top-performing annotator received an additional USD 100, while the second-highest received USD 35.\footnote{The annotation guidelines can be found at Appendix~\ref{app:annotation}}

\section{Results and Analysis}
\subsection{Prediction Accuracy}
Table~\ref{tab:accuracy} presents the evaluation results. Firstly, we observe significant differences in task difficulty between predicting intraday stock movements based on morning briefs and forecasting overnight market movements using closing-bell reports. The latter scenario appears inherently easier, as evidenced by consistently higher accuracy rates across all investors and information sources. Secondly, examining the morning brief scenario more closely, we find that decision accuracy based on original journalist-produced texts does not achieve optimal performance for either human or LLM investors. In contrast, LLM-generated morning briefs consistently lead to improved decision accuracy. This suggests that although traditional journalistic content is valuable, the analytical synthesis provided by LLM-generated summaries offers clearer signals or more actionable insights for immediate investment decisions. Thirdly, a detailed analysis reveals that incorporating expert-driven asset selection (professional-insight) consistently enhances predictive accuracy for both human investors and LLM agents. This finding shows the continuing value of human expertise at the initial stage of asset filtering. Notably, while performance-based selection can also yield beneficial outcomes, its effectiveness varies significantly depending on the specific investor, suggesting that human insight remains a uniquely stabilizing factor.

Turning to the closing-bell reports, we observe a notable difference in utility between human and LLM investors. Journalist-produced closing-bell reports substantially benefit LLM agents, who achieve the highest accuracy when utilizing these professionally authored texts. However, the opposite trend emerges for human investors, who consistently perform better when decisions are informed by LLM-generated reports. This divergent outcome prompts reconsideration of traditional evaluation metrics: Should the ultimate goal be lexical and content alignment with human-generated expert texts, or should we instead assess the practical decision-making value that LLM-generated texts uniquely provide? Our experimental results strongly support the latter perspective, suggesting that LLM-generated content should be evaluated primarily for its practical efficacy rather than its fidelity to existing human-authored texts. Finally, regarding asset selection methods, the professional-insight approach typically leads to superior decision-making outcomes across scenarios, reaffirming the importance of human expertise in the content generation pipeline. Although exact replication of expert commentary may be unnecessary, human involvement in guiding the asset selection and content framing process significantly enhances decision quality.

\begin{table}[t]
  \centering
  \resizebox{\columnwidth}{!}{
    \begin{tabular}{ll|r|r|r|r|r|r}
    \multicolumn{1}{c}{\multirow{2}[1]{*}{Investor}} & \multicolumn{1}{c|}{\multirow{2}[1]{*}{Decision}} & \multicolumn{3}{c|}{Morning Briefs} & \multicolumn{3}{c}{Closing-Bell Reports} \\
          &       & \multicolumn{1}{c|}{Journalist} & \multicolumn{1}{c}{PB} & \multicolumn{1}{c|}{PI} & \multicolumn{1}{c|}{Journalist} & \multicolumn{1}{c}{PB} & \multicolumn{1}{c}{PI} \\
    \hline
    \multirow{2}[2]{*}{LLM} & Buy   & 13.24 & 3.89  & 7.12  & 6.21  & 3.86  & 3.96 \\
          & Sell  & 4.17  & 1.19  & 1.30  & 2.37  & 2.06  & 1.28 \\
    \hline
    \multirow{2}[1]{*}{Human} & Buy   & 2.69  & 2.17  & 2.63  & 2.42  & 2.60  & 2.69 \\
          & Sell  & 1.36  & 0.56  & 0.96  & 1.22  & 1.33  & 1.01 \\
    \end{tabular}%
    }
  \caption{\textbf{Average number of transactions.} PB and PI denote performance-based and professional-insight.}
  \label{tab:Decision analysis}%
\vspace{-0.4cm}
  
\end{table}%

\subsection{Investor Behavior}
Table~\ref{tab:Decision analysis} further analyzes investor decision-making behavior by presenting the average number of transactions under various market digest scenarios. Across all conditions, both human and LLM investors consistently executed more buy than sell decisions. Notably, LLM investors engaged in a higher overall number of transactions than human investors, especially when relying on journalist-produced market digests. This may be attributed to the nature of traditional financial journalism, which often references a wide array of assets and market signals without providing clear directional emphasis. Such ambiguity may prompt LLMs to interpret multiple weak or conflicting cues as actionable, resulting in a higher frequency of trades.

By contrast, LLM-generated market digests, particularly morning briefs, tend to present more distilled and prioritized information, which reduces signal ambiguity and discourages excessive trading. This effect is reflected in the overall reduction of transactions for both human and LLM investors when using LLM-generated content. In particular, the number of sell decisions declined notably, with human investors exhibiting a marked decrease in selling activity when guided by LLM-produced morning briefs. One possible explanation is that LLM-generated texts may be implicitly biased toward highlighting positive developments or investment opportunities. This tendency likely results from the model’s training on general language data, which favors coherence, optimism, and recognizable patterns over uncertainty or negative sentiment. Consequently, this linguistic framing may influence both human and LLM investors to adopt a more optimistic outlook, increasing the likelihood of buy decisions while reducing sell-side activity.

\section{Toward Decision-Oriented Evaluation}

The variability observed in human investor performance presents significant challenges for reliably evaluating generated texts. Human decision-making is inherently influenced by personal biases, varying levels of domain expertise, and risk tolerance, which introduce substantial noise and complicate reproducibility and comparative analysis across different studies. Consequently, evaluating the efficacy of generated texts through human investors alone may yield inconsistent and difficult-to-generalize results.

In contrast, leveraging LLM investors offers distinct advantages for achieving consistent and reproducible evaluation outcomes. Under controlled settings (e.g., temperature set to zero), identical LLM models produce deterministic outputs, ensuring stable and replicable decision processes. By clearly documenting the model architectures, parameters, and input configurations, future research can precisely reproduce experimental conditions, thereby facilitating direct comparisons across studies and systematically assessing the impact of textual variations on decision-making efficacy.

Furthermore, this methodological framework extends beyond the use of contemporary LLMs for text evaluation, accommodating various text-based decision algorithms already developed within the computational finance and artificial intelligence communities. This flexibility allows for broader experimentation and benchmarking of different analytical approaches, enhancing our ability to rigorously evaluate and optimize textual content for decision support.
In sum, adopting LLM investors as standardized evaluators can significantly enhance the reliability, reproducibility, and comparability of decision-oriented text evaluation, providing a robust foundation for future research.

\section*{Limitation}
 First, our study's experimental design utilized short-term investment decisions based on intraday or overnight stock movements, potentially limiting the generalizability of findings to longer investment horizons or broader market contexts. Real-world financial decisions often incorporate long-term strategic considerations, risk management, and broader macroeconomic factors not fully captured in our short-term evaluation scenario.

Second, the LLM investor agents in our study, although deterministic under controlled conditions, inherently possess distinct analytical capabilities and biases reflective of their training data. Their decision-making processes may not fully emulate the complexity or adaptability of human investors, particularly experienced financial professionals who may integrate nuanced qualitative judgments or intuition beyond the information provided in textual summaries alone.

Third, the study focused exclusively on market digest texts related to financial markets, specifically morning briefs and closing-bell reports. This focus limits the direct applicability of our conclusions to other high-stakes domains such as medicine or law, which may involve significantly different decision-making criteria, textual characteristics, and consequences.

Finally, our evaluation framework emphasizes immediate decision accuracy as the primary performance metric. While accuracy is a critical and directly measurable outcome, it does not fully encapsulate other dimensions of decision quality, such as risk-adjusted returns, portfolio diversity, or ethical considerations in decision-making processes. Future research should integrate more comprehensive decision metrics to further validate and enhance the practical utility of generated texts in real-world settings.

\bibliography{ref}

\appendix

\begin{table*}[t]
  \centering
  \resizebox{\textwidth}{!}{
    \begin{tabular}{ll|rr|rr|rr|rr}
    \multicolumn{2}{c|}{Generator} & \multicolumn{4}{c|}{Gemini-2.0 Flash} & \multicolumn{4}{c}{GPT-4o} \\
    \hline
    \multicolumn{2}{c|}{\multirow{2}[2]{*}{Investor}} & \multicolumn{2}{c|}{Morning Brief} & \multicolumn{2}{c|}{Closing-Bells} & \multicolumn{2}{c|}{Morning Brief} & \multicolumn{2}{c}{Closing-Bells} \\
    \multicolumn{2}{c|}{} & \multicolumn{1}{c}{PB} & \multicolumn{1}{c|}{PI} & \multicolumn{1}{c}{PB} & \multicolumn{1}{c|}{PI} & \multicolumn{1}{c}{PB} & \multicolumn{1}{c|}{PI} & \multicolumn{1}{c}{PB} & \multicolumn{1}{c}{PI} \\
    \hline
    \multirow{3}[1]{*}{LLM} & Claude-3-5-Sonnet & 48.36\% & \textbf{51.09\%} & 48.44\% & 56.56\% & 45.98\% & 48.01\% & 55.62\% & \textbf{60.07\%} \\
          & Gemini-2.0-Flash & 43.39\% & \textbf{47.16\%} & 49.07\% & 54.92\% & 46.98\% & 42.41\% & 54.36\% & \textbf{58.25\%} \\
          & GPT-4o & \textbf{44.36\%} & 43.42\% & 47.57\% & 55.78\% & 42.53\% & 43.15\% & \textbf{56.89\%} & 55.17\% \\
    \end{tabular}%
    }
  \caption{Using Different LLMs for Generating Market Digest}
  \label{tab:Using Different LLMs for Generating Market Digest}%
\end{table*}%

\begin{table*}[t]
\centering
\begin{tabular}{p{0.45\linewidth} p{0.45\linewidth}}
\toprule
\textbf{Morning Brief} & \textbf{Closing Bell} \\
\midrule
Last Friday, U.S. markets closed higher—led by a 2.6\% surge in the Philadelphia Semiconductor Index—after President Trump’s weekend press conference omitted new China sanctions and trade‐deal terminations, focusing instead on targeted visa bans and student restrictions. 
U.S. futures later pulled back as nationwide protests over a fatal police shooting in Minnesota rekindled racial tensions. 
Meanwhile, Huawei’s August ban has spurred Chinese firms to accelerate chip self‐production, Hong Kong’s dollar remained freely convertible, and foreign investors sold NT\$11 billion of Taiwan stocks.  
Today, investors await the U.S. ISM manufacturing index and Friday’s non‐farm payrolls, which may shape tomorrow’s open \dots
&  
At today’s close, the Taiwan Stock Exchange rose 136 points to 11,079 on NT\$166 billion in volume.  
Foreign investors returned, lifting electronics to 68\% of turnover and driving the new market leader to NT\$4,065 on strong volume.  
Analysts note that June’s focus will be on companies with robust May revenues and the upcoming government voucher rollout boosting consumer sectors.  
Looking ahead, the electronics sector—anchored by TSMC and major chip designers—may sustain momentum if global reopening continues, while next week’s earnings and policy announcements will guide the market’s direction \dots
\\
\bottomrule
\end{tabular}
\caption{Illustrative examples of market commentary}
\label{tab:commentary_examples}
\end{table*}

\begin{table}[t]
\small
  \centering
  \begin{tabular}{lccc}
    \toprule
    Annotator &  A &  B &  C \\
    \midrule
    \makecell[l]{Working \\ Industry}              &  \makecell{Wholesale \\ \& Retail}     & \makecell{Information \\ Technology} & \makecell{Financial \\ Services} \\
    \hline
    \makecell[l]{Investing \\ Experience} & 2 years           & 4 years           & 7 years          \\
    \hline
    \makecell[l]{Risk \\ Aversion}         & Moderate             & Moderate             & Aggressive           \\
    \hline
    \makecell[l]{Investing \\ Budget}      & Low                  & Medium               & High                 \\
    \bottomrule
  \end{tabular}
  \caption{Annotator background summary}
  \label{tab:annotator-background}
\end{table}

\section{Comparison of Using Different LLMs for Generating Market Digest}
In the main text, we primarily focused our analyses and discussions on texts generated by GPT-4o. To justify this decision, we conducted additional experiments using another large language model, Gemini-2.0-Flash, across an extended dataset covering 89 days (approximately four months‘ trading days). Specifically, we generated market digests (morning briefs and closing-bell reports) using both Gemini-2.0-Flash and GPT-4o, then evaluated these digests using identical large language models (Claude-3-5-Sonnet, Gemini-2.0-Flash, and GPT-4o) as autonomous investors.
Table~\ref{tab:Using Different LLMs for Generating Market Digest} summarizes these comparative results.

\section{Example}\label{app:example}
Table~\ref{tab:commentary_examples} presents examples of morning briefings and closing-bell commentaries.

\section{Background of Annotators}
We asked the annotators to complete forms typically used by professional financial institutions to assess clients' investment risks. Table~\ref{tab:annotator-background} presents the annotators’ background information, including their professional experience, investment background, risk tolerance, and the size of their investable assets.

\section{Annotation Guidelines}
\label{app:annotation}

In this task, you will play the role of a Taiwan stock investor and make buy/sell decisions based on market news summaries. We hope to leverage your insights to capture subtle market shifts and compare your performance with that of large language models (LLMs).

\subsection{Rewards and Incentives}
You will be compensated at a base rate equal to 130\% of the minimum hourly wage for each hour of annotation. If your decisions outperform the average performance of the participating LLMs, you will earn an additional USD 65. We have also prepared ranking bonuses among the three annotators: the first-place annotator will receive an extra USD 100, and the second-place annotator will receive an extra USD 35. 

\subsection{Annotation Workflow}
\begin{enumerate}
  \item \textbf{Read the Summary:} Carefully read one financial market news summary and fully understand the information.
  \item \textbf{Adopt an Investor Mindset:} Imagine yourself as an active Taiwan stock investor, and consider what this summary implies for your portfolio.
  \item \textbf{Make Your Decisions:} In the annotation interface, you will see two multi-select lists containing all Taiwan-listed stocks. You may select any number of stocks:
    \begin{itemize}
      \item \textbf{Buy:} Select all stocks you predict will rise.
      \item \textbf{Sell:} Select all stocks you predict will fall.
      \item \textbf{No Action:} If the information is insufficient or you decide no trade is necessary, leave both lists empty and provide a brief remark (e.g., “Information too vague,” “No directly relevant stocks”).
    \end{itemize}
    You are not limited to companies explicitly mentioned in the summary. Feel free to apply industry insights, economic trends, and your own investment experience and risk preferences to related firms.
  \item \textbf{Remarks:} In the remarks field, provide a concise explanation of your overall portfolio strategy. Highlight only the most important points or the stocks you are most confident about, without detailing every selection.
\end{enumerate}

\subsection{Important Notes}
\begin{itemize}
  \item Do \emph{not} consult real historical data or news articles---this would be considered cheating.
  \item Do \emph{not} replicate actual past portfolios, as this would invalidate our evaluation.
  \item News summaries may be generated by humans or LLMs; do \emph{not} let assumptions about the source influence your decisions.
  \item Multiple summaries may describe the same day’s market open or close. Do \emph{not} assume they refer to the same timestamp---make decisions based solely on the information presented.
\end{itemize}

\subsection{Performance Evaluation}
\begin{itemize}
  \item \textbf{Open Summary:} Accuracy is measured against the same day’s closing price.
  \item \textbf{Close Summary:} Accuracy is measured against the next day’s opening price.
  \item A “Buy” decision is counted as correct if the stock rises by more than 0.55\%.
  \item A “Sell” decision is counted as correct if the stock falls by more than 0.50\%.
  \item All other outcomes are considered incorrect.
  \item Your success rate will be calculated by comparing your selections with actual stock movements.
\end{itemize}

Please refer to the provided \texttt{companies.csv} file for a full list of listed stocks and their codes, which you can use in the annotation interface.

\end{document}